JOURNAL OF COMPUTING, VOLUME 2, ISSUE 3, MARCH 2010, ISSN 2151-9617
HTTPS://SITES.GOOGLE.COM/SITE/JOURNALOFCOMPUTING/166# Land-cover Classification and Mapping for Eastern Himalayan State Sikkim

Ratika Pradhan, Mohan P. Pradhan, Ashish Bhusan, Ronak. K. Pradhan, M. K. Ghose

**Abstract**— Area of classifying satellite imagery has become a challenging task in current era where there is tremendous growth in settlement i.e. construction of buildings, roads, bridges, dam etc. This paper suggests an improvised k-means and Artificial Neural Network (ANN) classifier for land-cover mapping of Eastern Himalayan state Sikkim. The improvised k-means algorithm shows satisfactory results compared to existing methods that includes k-Nearest Neighbor and maximum likelihood classifier. The strength of the Artificial Neural Network (ANN) classifier lies in the fact that they are fast and have good recognition rate and it's capability of self-learning compared to other classification algorithms has made it widely accepted. Classifier based on ANN shows satisfactory and accurate result in comparison with the classical method.

**Index Terms**— Image classification, k-means, Artificial Neural Networks (ANN), perceptrons, signature sets.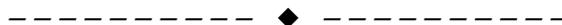

## 1 INTRODUCTION

The interpretation of remotely sensed data uses techniques from a number of disciplines including remote sensing, pattern recognition, artificial intelligence, computer vision, image processing and statistical analysis. The move towards automated analysis of remotely sensed data is encouraged by the ever increasing volumes of data as well as by the high cost of ground surveying. A number of methodologies have been developed and employed for image classification from remotely sensed data within the past 20 years. Classification is a method by which labels are attached to pixels in view of their character. This character is generally their response in different spectral ranges. Image classification techniques in remote sensing can be divided into supervised and unsupervised methods based on the involvement of the user during the classification process. Supervised classification techniques require training areas to be defined by the analyst in order to determine the characteristics of each category. Each pixel in the image is, thus, assigned to one of the categories using the extracted discriminating information. Unsupervised classification, on the other hand, searches for natural groups of pixels, called clusters, present within the data by means of assessing the relative locations of the pixels in the feature space. In these classification systems, an algorithm is used to identify unique clusters of points in feature space, which are then assumed to represent unique categories. Supervised learning is the more useful technique when the data samples have known outcomes that the user wants to predict. On the other hand, unsupervised learning is more appropriate when the user does not know the subdivisions into which the data samples should be divided. When ground information concerning the characteristics of individual classes is not available in land cover classification problems, an unsupervised classification technique is used to identify a number of distinct or separable categories. In other words, an unsupervised classification method is used to determine the number of spectrally-separable groups or clusters in an image for which there is insufficient ground reference information available. While applying an unsupervised method, the analyst generally specifies only the number of classes (or the upper and lower bound on the number of classes) and some statistical measure, depending upon the type of clustering algorithms used. These methods generate the specified number of clusters in feature space, and the user assigns these clusters (spectral classes) to information classes depending on his or her knowledge of the area. Determination of the clusters is performed by estimating the distances or comparison of the variance within and between the clusters. These automated classification methods are expected to delineate (or extract) those land cover features that are desired by the analyst. After the specified number of groups is determined, they are labelled by allocating pixels to land cover features present in the scene.

Supervised classification may be defined as the process of identifying unknown objects by using the spectral information derived from training data provided by the analyst. The result of the identification is the assignment of unknown pixels to pre-defined categories. The main difference between the unsupervised and supervised classification approaches is that supervised classification requires training data. The analyst locates specific sites in the remotely sensed image that represent homogeneous examples of known land cover types. These areas are commonly referred to as training sites because

————————————————
- *Ratika Pradhan is with the Department of Computer Sc. and Engg., Sikkim Manipal Institute of Technology, Sikkim, INDIA.*
- *Mohan P. Pradhan is with the Department of Computer Sc. and Engg., Sikkim Manipal Institute of Technology, Sikkim, INDIA.*
- *Ashish Bhusan is with the Department of Computer Sc. and Engg., Sikkim Manipal Institute of Technology, Sikkim, INDIA.*
- *Ronak. K. Pradhan is with the Department of Computer Sc. and Engg., Sikkim Manipal Institute of Technology, Sikkim, INDIA.*
- *M. K. Ghose is with the Department of Computer Sc. and Engg., Sikkim Manipal Institute of Technology, Sikkim, INDIA.*

the spectral characteristics of these known areas are used to train the classifier. The training data thus extracted is used to find the properties of each individual class.

Supervised classification is performed in two stages; the first stage is the training of the classifier, and the second stage is testing the performance of the trained classifier on unknown pixels. In the training stage, the analyst defines the regions that will be used to extract training data, from which statistical estimates of the data properties are computed. At the classification stage, every unknown pixel in the test image is labelled in terms of its spectral similarity to specified land cover features. If a pixel is not spectrally similar to any of the classes, then it can be allocated to an unknown class. As a result, an output image, or thematic map is produced, showing every pixel with a class label. The characteristics of the training data selected by the analyst have a considerable effect on the reliability and the performance of a supervised classification process. The training data must be defined by the analyst in such a way that they accurately represent the characteristics of each individual feature and class used in the analysis. Supervised classification methods require more user interaction, especially in the collection of training data. The accuracy of supervised classification is determined partly by the quality of the ground truth data and partly by how well the set of ground truth pixels are representative of the full image.

Statistical image classification techniques are ideally suited for data in which the distribution of the data within each of the classes can be assumed to follow a theoretical model. The most commonly used statistical classification methodology is based on nearest neighbour, a pixel-based statistical classification method which assumes that spectral classes can be described by a statistical distribution in multi-spectral space. This traditional approach to classification is found to have some limitations in resolving interclass confusion if the data used are not normally distributed. As a result, in recent years, and following advances in computer technology, alternative classification strategies have been proposed. In most instances, human beings are good pattern recognisers. This observation led researchers in the field of pattern recognition to consider whether computer systems based on a simplified model of the human brain can be more effective than the standard statistical and knowledge-based classification methods. Research in this field led to the adoption of artificial neural networks (ANN), which have been used in remote sensing over the past ten years, mainly for image classification. Studies carried out using ANN suggest that, due to their nonparametric nature, they generally perform better than statistical classifiers. The performance of a neural network classifier depends to a significant extent on how well it has been trained. During the training phase, the neural network learns about regularities present in the training data and, based on these regularities, constructs rules that can be extended to the unknown data. However, the user must determine a number of properties such as the architecture of network, learning rate, number of iterations and learning algorithms, all of which affect classification accuracy. There is no clear rule to fix the values of these parameters, and only rules of thumb exist to guide users in their choice of network parameters.

## 2 RELATED WORK

There exist various classification methods for classifying satellite imagery which includes Maximum likelihood classifier, k-NN classifier, k-means classifier, parallel piped classifier and fuzzy classifier. Multilayer perceptrons have been applied to variety of problems in image processing, including optical character recognition [7] and medical diagnosis [3, 4]. A. Tzotsos and D. Argialas, has suggested Support Vector Machine classification technique for Object Based Image Analysis (OBIA) for supervised classification of the satellite imagery using object based representation [1]. X. Gigandet et.al. have suggested region based satellite image classification that combines unsupervised segmentation with supervised segmentation using Gaussian hidden Markov random field and Support Vector Machine [8].

## 3 STUDY AREA

Sikkim is 22nd state of India and is mountainous state that shares its boundary with three sovereign nations, Nepal in west, Bhutan in east and Tibet, China in north. The state shares its southern boundary with Darjeeling district of West Bengal, India. The state is situated between $27^0 04'46''$ and $28^0 07'48''$ north latitudes and $88^0 00'58''$ and $88^0 55'25''$ east longitude. The state has total geographical area of 7096 sq km. Sikkim has a very rugged topography and formidable physical feature. Northern region of Sikkim is mostly covered by the snow and has no populated area except Lachen and Lachung. Southern Sikkim is densely populated and is fairly cultivated in patches. It is subjected to erosion by River Teesta and its tributaries.

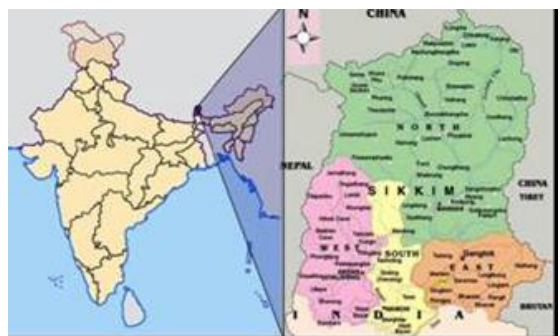

Figure 1: Location of State Sikkim

## 3 DATA & SOFTWARE USED

We have used LISS III multi-spectral data for the State Sikkim with 24m resolution. The satellite imagery was first geo-referenced with the help of ERDAS IMAGINE 9.3 Software and then result for unsupervised classification was obtained using ERDAS IMAGINE 9.3 to compare it with the result obtained using ANN classifier for both supervised and unsupervised classification methods.

For ANN classifier to design and implement, we have used MATLAB 7.

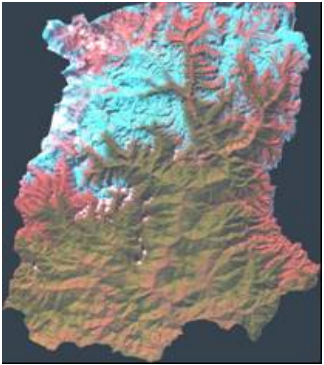

Figure 2: Multi-spectral LISS II Satellite Imagery for State Sikkim

## 4 IMPROVISED K-MEANS CLASSIFICATION

k-means classifier is widely used for classifying satellite imagery but it is biased on initial mean value selected and it tends to misclassify the pixel value to different class. Here we have tried to improvise the k-means using the algorithm given below:

i. Calculate the mean of each row of the entire image.

$$\bar{x} = \frac{1}{N}\sum_{i=1}^{N} x_i \qquad (1)$$

where N is the number of pixels in each row.

ii. Find the minimum and maximum of the mean of the entire image.

iii. Depending on the number of classes define the range of variation for each of the classes as

$range = (\max - \min)/k$ (2)

where k is the number of classes.

iv. Calculate distance metric between pixels and the mean value.

$$d_e(x,y) = \sqrt{\sum_{i=1}^{N}(x_i - y_i)^2} \qquad (3)$$

$$\sigma(x) = \sqrt{\frac{1}{N}\sum_{i=1}^{N}(x_i - \bar{x})^2} \qquad (4)$$

v. Classify all pixels to respective classes based on the distance metric and the range of means.

vi. When all the pixels will be allotted their respective classes then the new mean for the classes will be calculated.

vii. Repeat the steps iv to vi throughout the image plane.

viii. Assign each class a unique combination of colors for each of the three layers.

## 5 IMAGE CLASSIFICATION USING NEURAL NETWORK

In order to classify satellite imagery, the first step is to preprocess the satellite imagery to suppress noise and then to normalize the input data i.e. set of pixel value. We have used Multilayer perceptrons (MLP) as the base neural network for the classification of the satellite imagery.

### 5.1 Multilayer Perceptron (MLP)

Typically the network consists of a set of processing units that constitute the input layer, one or more hidden layers, and an output layer. The input signal propagates through the network in forward direction, on layer- by-layer basis. The weigh $w_{ij}$ connects input node $x_i$ to hidden node $h_j$ and weight $v_{jk}$ connects hidden node $h_j$ to output node $o_k$. Classification begins by presenting a pattern to the input nodes $x_i$, $1 \le i \le l$. The data flow in one direction through the perceptron until the output nodes $o_k$, $1 \le k \le n$ are reached. Output nodes will have the value either 0 or 1. Thus, the perceptron is capable of portioning its pattern space into $2^n$ classes. The following steps are used for classification [6]:

i. Present pattern $\mathbf{p} = [p_1, p_2, \ldots, p_l] \in R^1, 1 \le i \le l$. to the perceptron, that is, set $x_i = p_i$ for $1 \le i \le l$.

ii. Compute the values of the hidden-layer nodes $h_j$

$$h_j = \frac{1}{1 + \exp\left[-(w_{oj} + \sum_{i=1}^{l} w_{ij}x_i)\right]} \quad 1 \le j \le m \qquad (5)$$

iii. Calculate the values of the output nodes according to

$$o_k = \frac{1}{1 + \exp\left[v_{jk} + \sum_{j=1}^{m} v_{jk}h_j\right]} \qquad (6)$$

iv. The class $c = [c_1, c_2, \ldots c_n]$ that the perceptron assigns p must be a binary vector. So $o_k$ must be the threshold of a certain class at some level τ.

v. Repeat steps i to iv for each pattern that is to be classified.

Multilayer perceptrons classification algorithm has been successfully applied for classification of the satellite imagery to identify the various classes in unsupervised mode. For supervised classification we have used back-propagation– type neural network where the network were trained initially by processing the training data set i.e. the *signature set*.

### 5.2 Back-propagation type neural network

Back-propagation process consists of two passes through the different layers of the network: a forward and a backward pass. During forward pass a training pattern is presented to the perceptron and classified. The backward pass recursively, level by level, determines error terms that are used later to adjust the perceptron weights. The error terms at the first level of the recursions are a function of $c^t$ and output of the perceptron $(o_1, o_2, \ldots, o_n)$. Once all the errors have been computed, weights are adjusted using the error terms that corresponds to their level. The following steps are carried out for back-propagation [6]:

i. Initialize the weights of the perceptron randomly with numbers between -0.1 and 0.1; that is,

 $w_{ij}$=random[-0.1,0.1] $0 \le i \le l$, $1 \le j \le m$

$w_{jk}$=random[-0.1,0.1] $0 \leq j \leq l, 1 \leq k \leq m$

ii. Present $p^t=[p_1^t, p_2^t,..p_l^t]$ from the training pair ($p^t,c^t$) to the perceptron and apply steps i to iii from the perceptron classification algorithm described earlier.

iii. Compute the errors $\delta_{ok}$, $1 \leq k \leq n$ in the output layer using

$$\delta_{ok} = o_k(1-o_k)(c_k^t - o_k) \qquad (7)$$

where $c^t = [c_1^t, c_2^t, ........, c_n^t]$ represents correct class of $p^t$.

The vector ($o_1, o_2, . . . , o_n$) represents the output of the perceptron.

Compute the errors $\delta_{hj}$, $1 \leq j \leq m$, in the hidden-layers nodes using

$$\delta_{hj} = h_j(1-h_j)\sum_{k=1}^{n}\delta_{ok}v_{jk} \qquad (8)$$

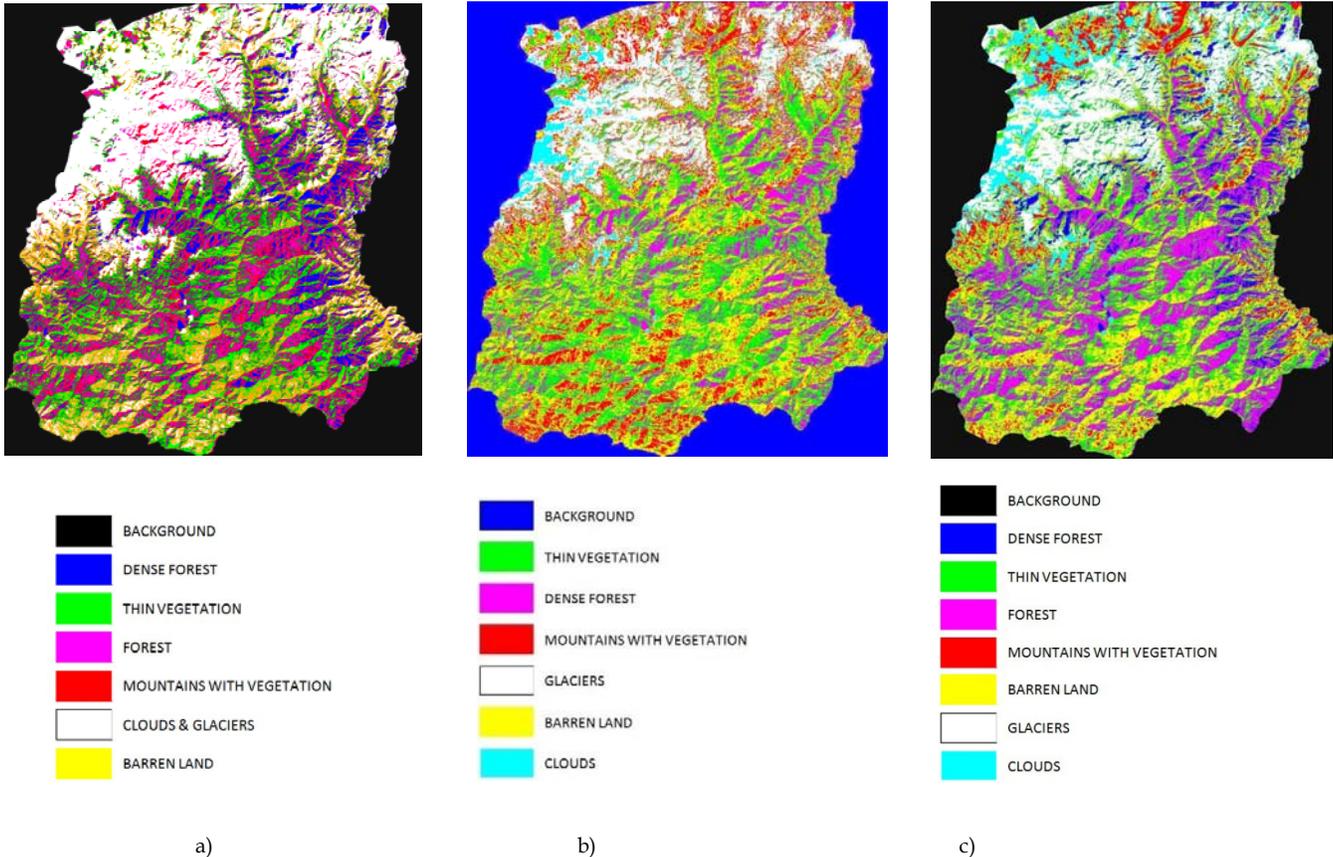

a)  b)  c)

Figure 3 a) Classified Image based on Improvised k-Means method b) Classified Image based on MLP Neural Network c) Classified image based on Back-propagation Neural Network

iv. Let $v_{jk}$ denote the value of weight $v_{jk}$ after the $t^{th}$ training pattern has been presented to the perceptron. Adjust the weights between the output layer and the hidden layer using

$$v_{jk}(t) = v_{jk}(t-1) + \eta\delta_{ok}h_j \qquad (9)$$

The parameter $0 \leq \eta \leq 1$ represents the learning rate. Adjust the weights between the hidden layer and the input layer according to

$$w_{ij}(t) = w_{ij}(t-1) + \eta\delta_{hj}p_i^t \qquad (10)$$

v. **Iteration:** Repeat steps 2 through 6 for each element of the training set. One cycle through the training set is called iteration.

From both performance and computational point of view, the size of the network is an important consideration. It has been shown [8] that one hidden layer is sufficient to approximate the mapping of any continuous function. In our wok therefore we have considered one hidden layer comprising of five neurons to reduce the danger of overtraining. The number of neurons in the input layer is equal to the number of the band available in the satellite imagery. And the number of nodes in the output layer is the number of classes or region on which the given satellite imagery is to be classified.

## 5 RESULTS & CONCLUSION

The LISS III image of the state Sikkim is classified using improvised k-Means, Unsupervised ANN using Multi-Layer Perceptron and Supervised ANN using Back-propagation Algorithm to train the neural network. The figure shown in 3(a) - (c) shows the result of applying these classification algorithms.

From the classified image it is evident that ANN has good recognition rate and computationally efficient compared to the improvised k-Means algorithms. Cloud re-gion are not classified in case of improvised k-Means as the pixel value for cloud is misclassified with that of the glacier. However, ANN is able to classify the cloud region more accurately. Another problem with improvised k-mean is that the shadow is often misclassified with that of background. We have come to the conclusion from the wok carried out that for the hilly region like Sikkim were we find very irregular surface pattern ANN can be used to classify the region correctly. For unsupervised classification method of classifying image observer is the best judge; however one can visit the place physically of ground truth collection. For supervised classification method the confusion matrix is laid down in the table below which gives overall classification accuracy of 90.70%.

From the confusion matrix shown in table 1 it is clear that supervised ANN gives best result for almost all the classes except for the thick forest region, as some the pixel values are mis-classified to the other forest classes i.e. thin

| Ground Classes | Background | Dense Forest | Thick Forest | Thin Vegetation | Barren Land | Glacier | Cloud | Omissions | Commissions | Map Accuracy |
|---|---|---|---|---|---|---|---|---|---|---|
| Background | 225 | 0 | 0 | 0 | 0 | 0 | 0 | 0% | 0% | 100% |
| Dense Forest | 0 | 220 | 0 | 0 | 0 | 5 | 0 | 2.22% | 5.77% | 92.44% |
| Thick Forest | 0 | 6 | 87 | 74 | 3 | 0 | 0 | 48.82% | 15.29% | 44.39% |
| Thin Vegetation | 0 | 0 | 19 | 206 | 0 | 0 | 0 | 8.44% | 34.22% | 68.21% |
| Barren Land | 0 | 0 | 3 | 3 | 422 | 3 | 0 | 2.08% | 0.69% | 97.23% |
| Glacier | 0 | 7 | 4 | 0 | 0 | 439 | 0 | 2.44% | 1.77% | 95.85% |
| Cloud | 0 | 0 | 0 | 0 | 0 | 0 | 225 | 0% | 0% | 100% |

Table1: Confusion matrix for supervised ANN classification algorithm.

vegetation and dense forest. In addition to the above classes it is also observed from the classified image that ANN has capability to segment the river compared to the other classification algorithms. Figure 4 shows number of pixels classified to various classes using various techniques.

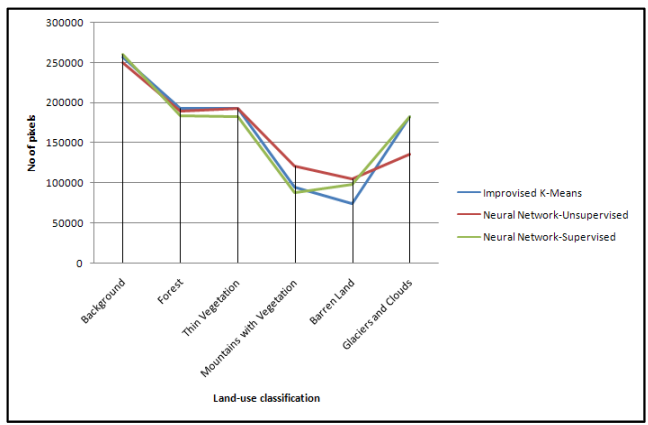

Figure 4 No. of pixel associated with various classes.

## 6 FUTURE SCOPE

The current work supports only one hidden layer and one input node from where the pixel value of the images is forwarded to the layer above it to move to any one of the classes. We can feed group of homogeneous pixel together through various input nodes to speed up the classification algorithms. The algorithm can be further improvised by considering region specific classification algorithms.